\documentclass[11pt]{article}

\usepackage[utf8]{inputenc}
\usepackage[T1]{fontenc}
\usepackage[margin=1in]{geometry}
\usepackage{amsmath,amssymb}
\usepackage{graphicx}
\usepackage{caption}
\usepackage{subcaption}
\usepackage{booktabs}
\usepackage{hyperref}
\usepackage{natbib}
\usepackage{authblk}

\title{Efficient Special Stain Classification}

\author[1,2,3]{Oskar Thaeter}
\author[1,2,3,4]{Christian Grashei}
\author[1]{Anette Haas}
\author[1]{Elisa Schmoeckel}
\author[1,2,3]{Han Li}
\author[1,2,3,4]{Peter Schüffler}
\affil[1]{Institute of Pathology, TUM School of Medicine and Health, Technical University of Munich, Germany}
\affil[2]{Department of Computer Science, TUM School of Computation, Information and Technology, Technical University of Munich, Germany}
\affil[3]{Munich Center for Machine Learning, Germany}
\affil[4]{Munich Data Science Institute, Technical University of Munich, Germany}

\date{}

\begin{document}

\maketitle

\begin{abstract}
Stains are essential in histopathology to visualize specific tissue characteristics, with Haematoxylin and Eosin (H\&E) serving as the clinical standard. However, pathologists frequently utilize a variety of special stains for the diagnosis of specific morphologies. Maintaining accurate metadata for these slides is critical for quality control in clinical archives and for the integrity of computational pathology datasets.
In this work, we compare two approaches for automated classification of stains using whole slide images, covering the 14 most commonly used special stains in our institute alongside standard and frozen-section H\&E. We evaluate a Multi-Instance Learning (MIL) pipeline and a proposed lightweight thumbnail-based approach.
On internal test data, MIL achieved the highest performance (macro F1: 0.941 for 16 classes; 0.969 for 14 merged classes), while the thumbnail approach remained competitive (0.897 and 0.953, respectively). On external TCGA data, the thumbnail model generalized best (weighted F1: 0.843 vs. 0.807 for MIL). The thumbnail approach also increased throughput by two orders of magnitude (5.635 vs. 0.018 slides/s for MIL with all patches).
We conclude that thumbnail-based classification provides a scalable and robust solution for routine visual quality control in digital pathology workflows.
\end{abstract}

\section{Introduction}

\begin{figure}[h]
\centering
\begin{subfigure}[b]{\linewidth}
    \centering
    \includegraphics[width=\linewidth]{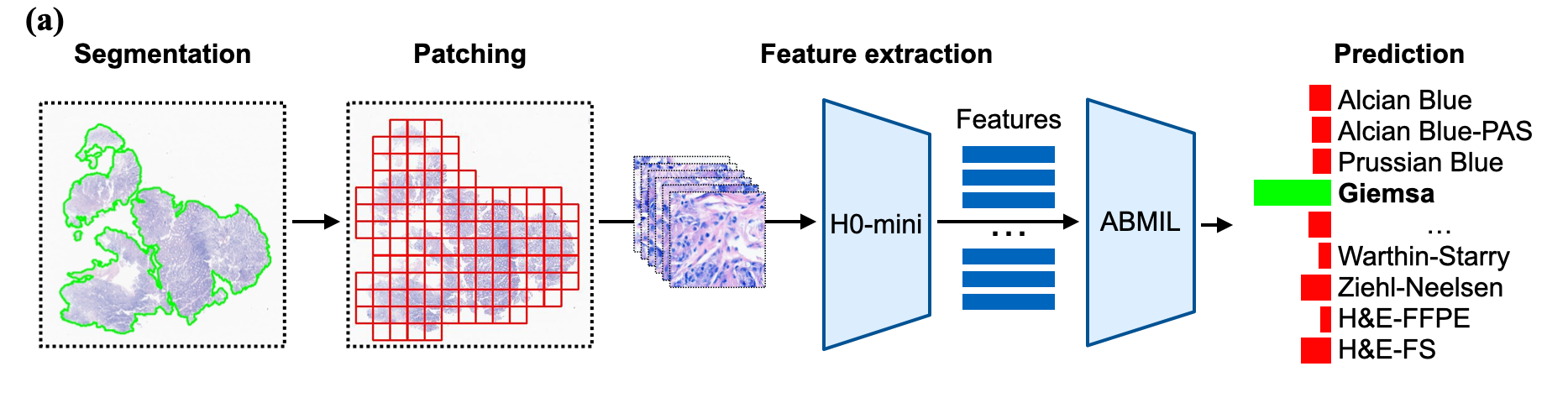}
    \caption{Standard MIL approach.}
    \label{fig:fig_1_mil}
\end{subfigure}

\vspace{1em}

\begin{subfigure}[b]{\linewidth}
    \centering
    \includegraphics[width=\linewidth]{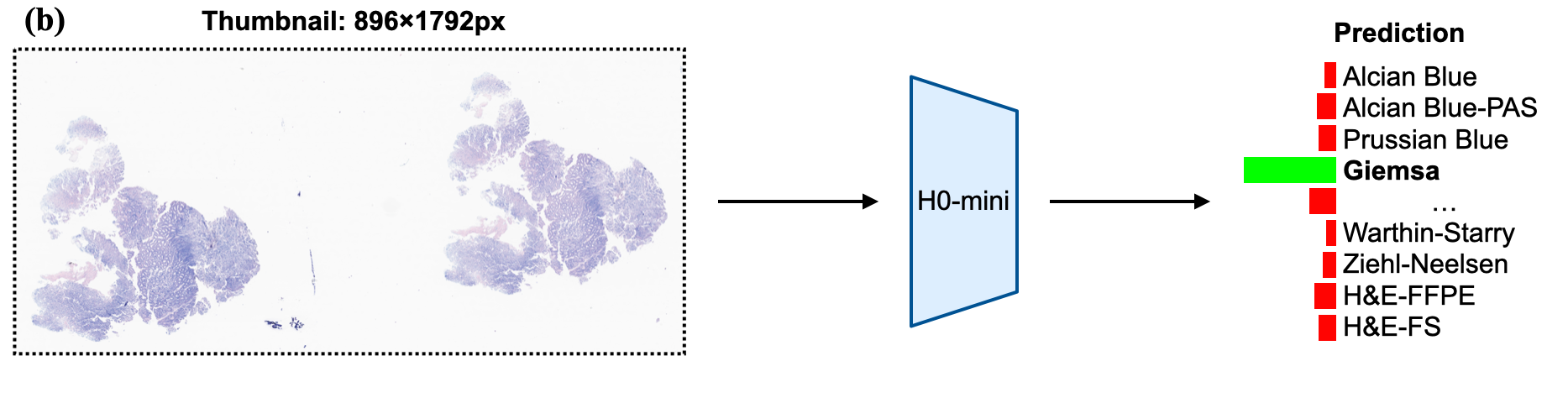}
    \caption{Proposed thumbnail approach.}
    \label{fig:fig_1_thumbnail}
\end{subfigure}

\caption{
Overview of the two classification pipelines.
(a)~The slide is segmented, divided into patches, and patch features are extracted and aggregated by ABMIL for prediction.
(b)~The whole slide thumbnail ($896 \times 1792$\,px) is processed in a single forward pass, avoiding tissue segmentation and patch-level feature extraction.
}
\label{fig:overview}
\end{figure}

Staining is a central step in histopathology, enabling pathologists to visualize tissue characteristics that are otherwise invisible to the naked eye.
While Haematoxylin and Eosin (H\&E) serves as the routine standard for the majority of tissue samples,
``special stains'' are frequently required to highlight particular structures, pathogens, or deposits essential for the diagnosis of specific morphologies.
Quality control in clinical laboratories must ensure the accuracy of slide annotations, including organ site, fixation type, and stain type.
Correct labelling is essential, as annotation errors can distort diagnosis by providing misleading visual cues.
Furthermore, in the era of computational pathology, mislabelled stains can silently corrupt large-scale datasets used for training deep learning models.

The manual inspection of metadata in large pathology datasets is often infeasible, allowing errors to persist undetected.
During the course of this study, we encountered multiple cases of incorrect or outdated labels within our own institute's archive, highlighting this issue in practice. In one extreme instance, we identified an H\&E-stained slide that was erroneously annotated as Ziehl-Neelsen stain. Such errors were often only discovered after a complete manual review, which required the recollection of cohorts and led to considerable delays.
These findings motivate the development of an efficient, automated visual classifier capable of flagging discordant slide labels before they impact clinical or computational workflows.

Previous stain classification works have focused on differentiating H\&E and Immunohistochemistry (IHC) stains\cite{Weng2019Metadata, yan_pathorchestra_2025}.
A larger body of work related to special stains focuses on generating images of different stains from H\&E stained images\cite{BCI2022, DeReStainer2024, UMDSTKidney2022}.

To the best of our knowledge, this work presents the first comprehensive effort to differentiate between multiple special stains, going beyond simple H\&E/IHC prediction.

In this work, we address the task of special stain classification covering the 14 most common special stains in routine use at our institute, namely Alcian Blue, Alcian Blue-PAS, Prussian Blue, Giemsa, Grocott's Methenamine Silver (GMS), Congo Red, Von Kossa, Rhodanine, Periodic acid-Schiff (PAS), PAS-Diastase, Reticulin, Van Gieson, Warthin-Starry and Ziehl-Neelsen, in addition to formalin-fixed, paraffin-embedded H\&E (H\&E-FFPE) and frozen-section H\&E (H\&E-FS). H\&E-FFPE and H\&E-FS are not distinct stains but different fixation methods. The fixation procedures cause large enough qualitative differences in the tissue as to warrant treating them as separate stains.
We compare two distinct computational approaches that offer different trade-offs between accuracy and computational efficiency (Figure~\ref{fig:overview}).
First, we evaluate a Multi-Instance Learning (MIL) pipeline, which aggregates features from all tissue patches in a slide\cite{MIL2001, ABMIL2018}.
While this patch-aggregation method yields high accuracy, it incurs a significant computational cost due to the necessity of tissue segmentation and exhaustive feature extraction. We therefore also test the same MIL pipeline with a fixed patch-budget, reducing the time needed for patch loading and feature extraction.
Alternatively, we propose a lightweight thumbnail-based approach that utilizes a single downscaled view of the whole slide.

We evaluated both approaches on an internal held-out test set and assessed H\&E-FFPE and H\&E-FS classification on external TCGA data.

\section{Results}

\subsection{Internal Validation on TUM Dataset}\label{subsection:TUM_results}

We evaluated the approaches on the internal, held-out test dataset. We report results for two class sets: fine which contains all 16 distinct stain classes, and coarse where closely related stain pairs (Alcian Blue/Alcian Blue-PAS, and PAS/PAS-D) were merged to form 14 classes.

\begin{table}[ht]
\centering
\small
\setlength{\tabcolsep}{4pt}
\begin{tabular}{l cc cc cc}
\toprule
& \multicolumn{2}{c}{\textbf{TUM fine (16)}} & \multicolumn{2}{c}{\textbf{TUM coarse (14)}} & \multicolumn{2}{c}{\textbf{TCGA (Ext.)}}\\
\cmidrule(lr){2-3} \cmidrule(lr){4-5} \cmidrule(lr){6-7}
\textbf{Approach} & F1 & W-F1 & F1 & W-F1 & F1 & W-F1 \\
\midrule
Thumbnail            & 0.897             & 0.907             & 0.953             & 0.965             & \textbf{0.562}       & \textbf{0.843}    \\[3pt]
MIL ($k{=}20$)       & 0.931$\pm$0.004   & 0.936$\pm$0.004   & 0.956$\pm$0.003   & 0.961$\pm$0.003   & 0.511$\pm$0.001      & 0.768$\pm$0.002   \\[3pt]
MIL ($k{=}\text{all}$) & \textbf{0.941}  & \textbf{0.949}    & \textbf{0.969}    & \textbf{0.978}    & 0.537                & 0.807             \\
\bottomrule
\end{tabular}
\caption{Comparison of classification performance using macro F1 score (F1) and weighted F1 score (W-F1) across approaches on internal (TUM) and external (TCGA) datasets. The ``coarse'' set merges closely related special stains. The best scores are highlighted in each column.}
\label{tab:results}
\end{table}

As shown in Table~\ref{tab:results}, the MIL approach achieved the highest performance on the internal dataset, reaching a macro F1 score of $0.941$ on the fine class set and $0.969$ on the coarse class set. With a fixed patch budget ($k=20$), the MIL configuration maintained high performance with macro F1 scores of $0.931$ and $0.956$ on the fine and coarse class sets, respectively. This demonstrates that a small subset of high-magnification patches is sufficient for accurate classification.

The thumbnail-based approach, while less accurate than the patch-aggregation methods on the fine set (F1: $0.897$), remained highly competitive on the coarse set (F1: $0.953$). The gap between MIL ($k=\text{all}$) and thumbnail shrank from $0.044$ on the fine set to $0.016$ on the coarse class set.

\begin{figure}[p]
\centering

\begin{subfigure}[b]{0.48\linewidth}
    \centering
    \includegraphics[width=\linewidth]{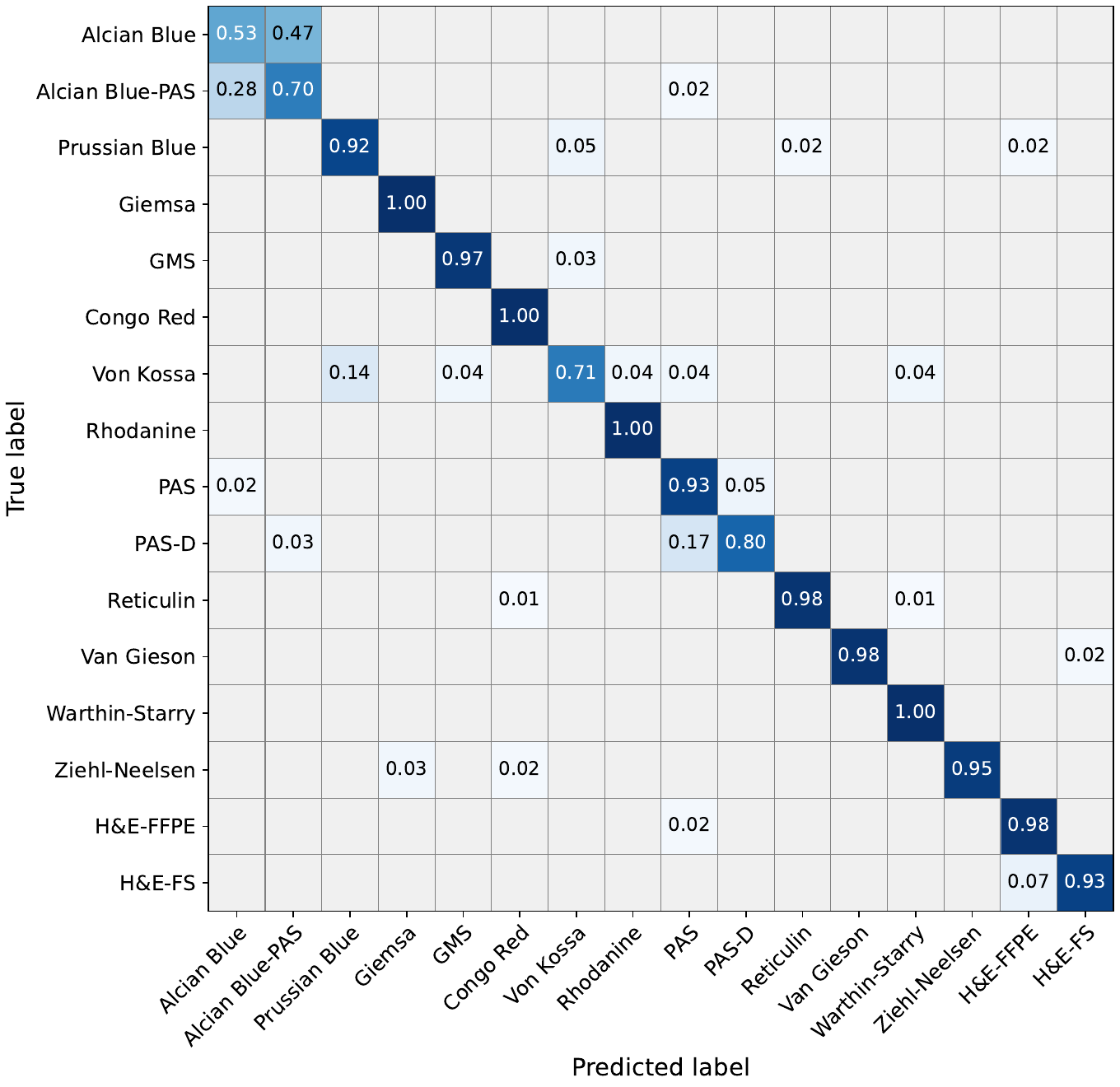}
    \caption{Thumbnail}
    \label{fig:thumbnail_fine_confusion}
\end{subfigure}
\hfill
\begin{subfigure}[b]{0.48\linewidth}
    \centering
    \includegraphics[width=\linewidth]{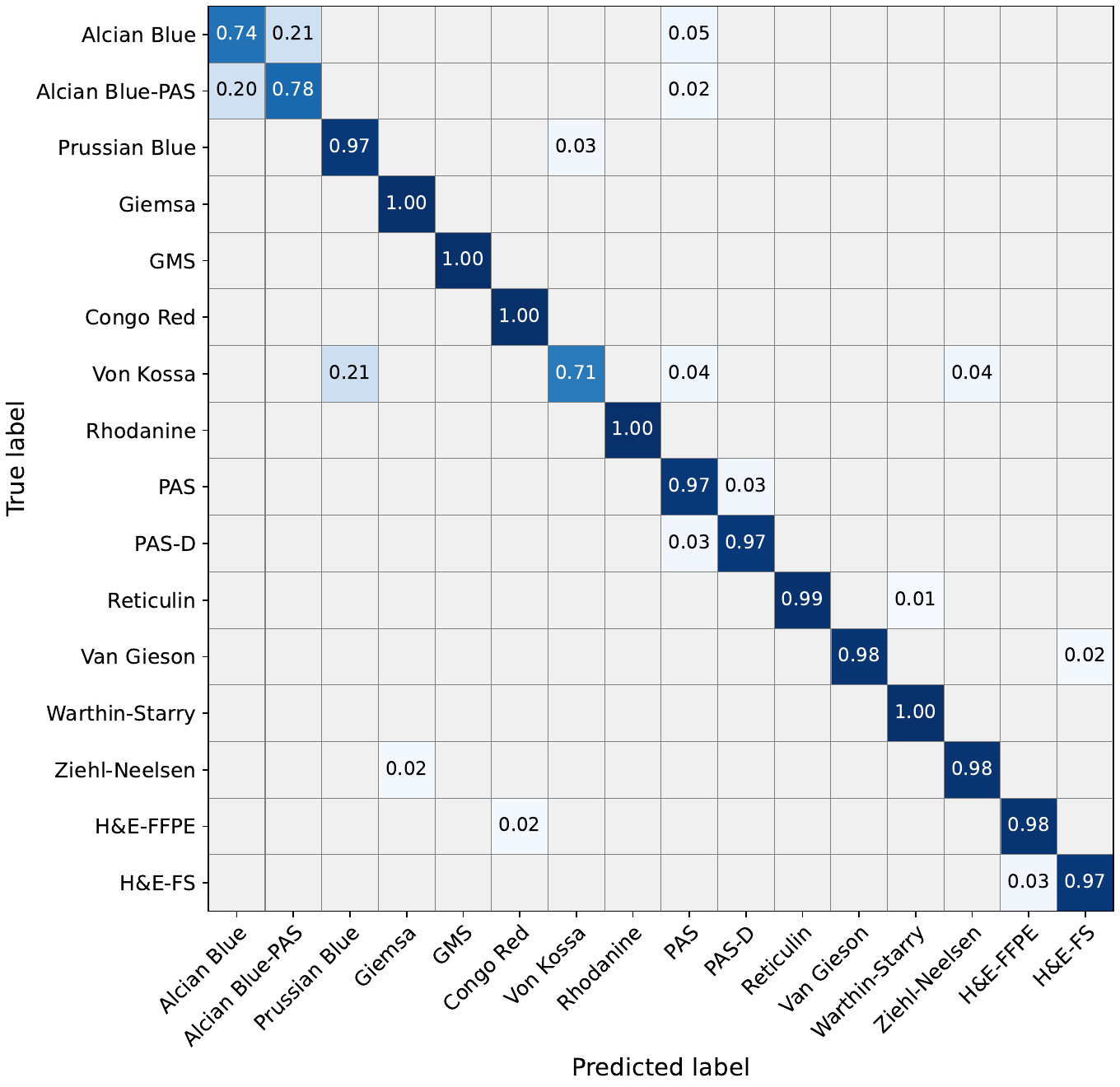}
    \caption{MIL ($k=\text{all}$)}
    \label{fig:mil_fine_confusion}
\end{subfigure}

\caption{Confusion matrices for thumbnail and MIL ($k=\text{all}$) approaches on the TUM test set (fine, 16 classes). Values below 0.01 are omitted for clarity.}
\label{fig:combined_confusions}
\end{figure}

The thumbnail approach struggles to differentiate Alcian Blue/Alcian Blue-PAS and PAS/PAS-D, as shown in Figure~\ref{fig:thumbnail_fine_confusion}. These shortcomings are resolved in the coarse class set, as these class pairings are merged.
MIL also has issues with differentiating Alcian Blue/Alcian Blue-PAS, although less severe (Figure~\ref{fig:mil_fine_confusion}). PAS/PAS-D differentiation is not an issue for MIL.

\begin{figure}[p]
\centering

\begin{subfigure}[b]{0.48\linewidth}
    \includegraphics[width=\linewidth]{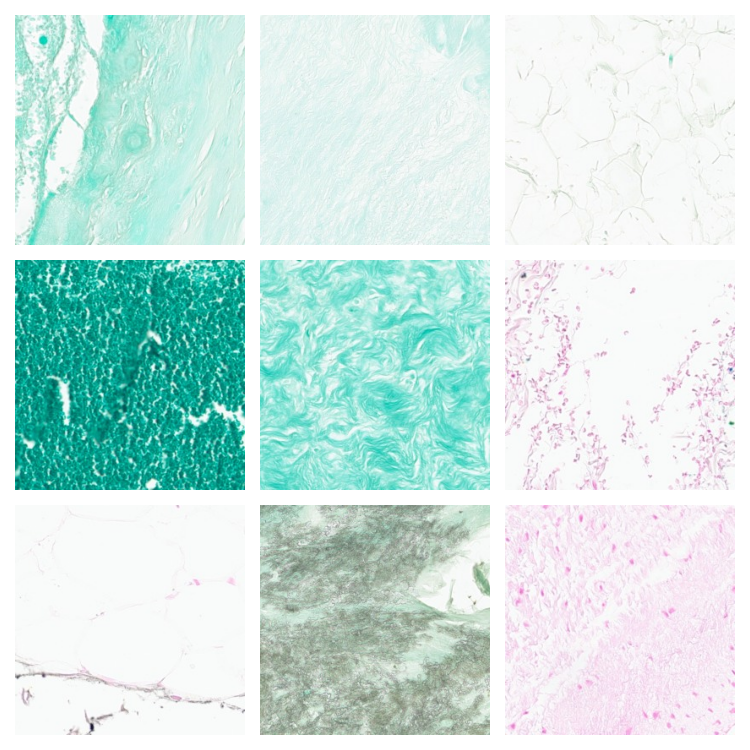}
    \caption{Von Kossa}
\end{subfigure}
\begin{subfigure}[b]{0.48\linewidth}
    \includegraphics[width=\linewidth]{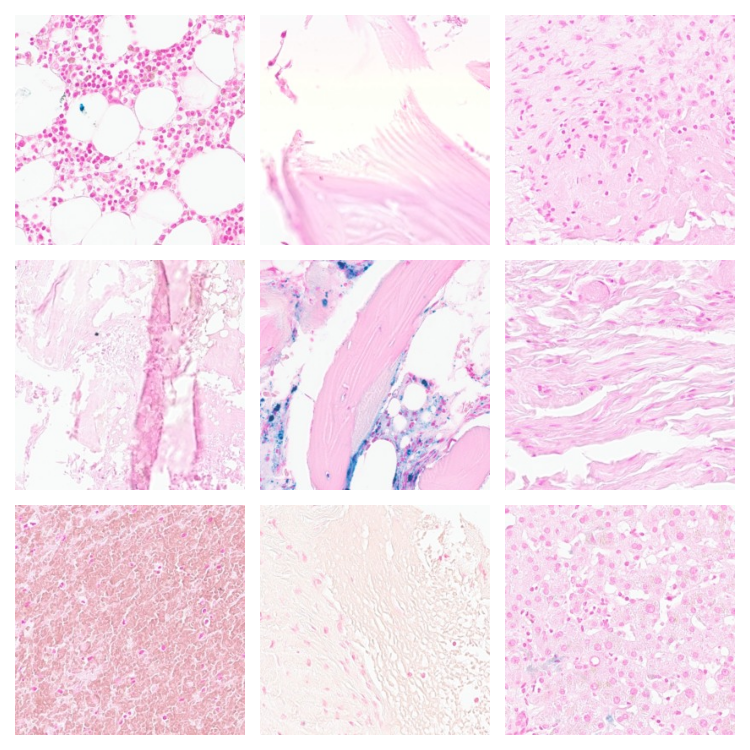}
    \caption{Prussian Blue}
\end{subfigure}

\caption{Representative $40\times$-magnification tiles of Von Kossa and Prussian Blue showing the sometimes used pink counterstain in Von Kossa, identical to the counterstain used in the Prussian Blue staining protocol.}
\label{fig:kossa_eisen_comparison}
\end{figure}

Both approaches struggle with the Von Kossa class. A sizeable proportion is misclassified as Prussian Blue. Looking at examples of both classes (Figure~\ref{fig:kossa_eisen_comparison}), we see Von Kossa sometimes uses the same pink counterstain as Prussian Blue.

\subsection{Generalization to External TCGA Data}\label{subsection:TCGA_results}
To assess model robustness, we evaluated the trained models on an external dataset from The Cancer Genome Atlas (TCGA),
which contains only H\&E-FFPE and Frozen Section H\&E (H\&E-FS) slides. Predictions for classes not represented in this test set were aggregated into a single ``Other'' class.

Contrary to the internal test results, the lightweight thumbnail approach demonstrated superior generalization, achieving a weighted F1 score of 0.843, surpassing both MIL configurations ($k=20$: $0.768$; $k=\text{all}$: $0.807$).

To further explore the subtask of fixation-type prediction, we trained each method only on H\&E-FFPE/H\&E-FS samples from the TUM training set to obtain a binary classifier.
In this subtask, the thumbnail approach achieved a macro F1 score of $0.885$ and an AUROC of $0.974$. This surpasses prior fixation-type prediction work using thumbnails alone (AUROC $0.88$\cite{thaeter2026}) and outperforms methods incorporating additional context including patch samples and free text metadata (AUROC $0.91$\cite{Weng2019Metadata}), while using only the thumbnail image as input. MIL ($k=\text{all}$) achieved a macro F1 score of $0.872$ and an AUROC of $0.984$.

\subsection{Attention Maps}\label{subsection:attention_maps}
\begin{figure}[h]
\centering
\begin{subfigure}[b]{0.49\linewidth}
    \centering
    \includegraphics[width=\linewidth]{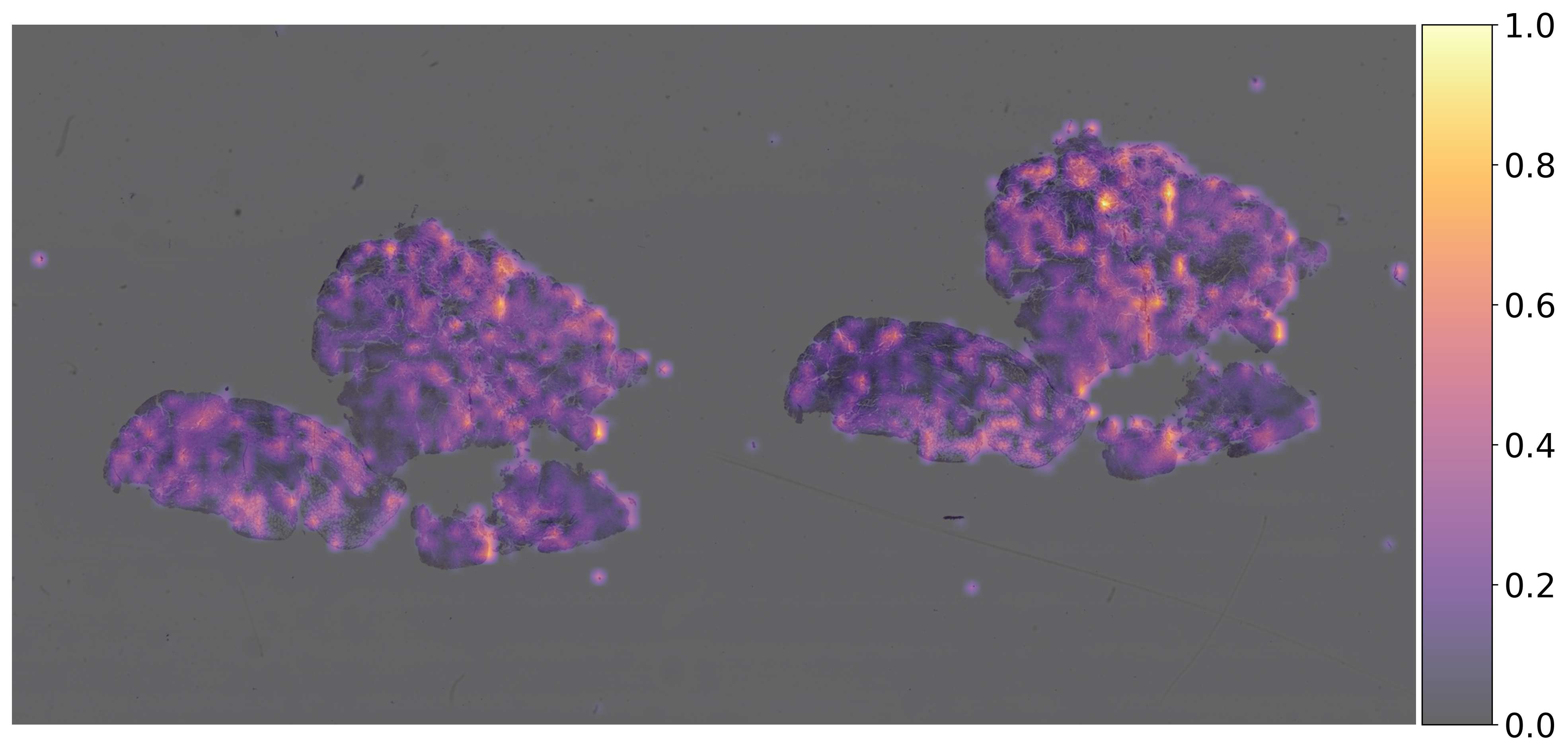}
    \caption{Thumbnail GradCAM}
    \label{fig:thumbnail_gradcam}
\end{subfigure}
\hfill
\begin{subfigure}[b]{0.49\linewidth}
    \centering
    \includegraphics[width=\linewidth]{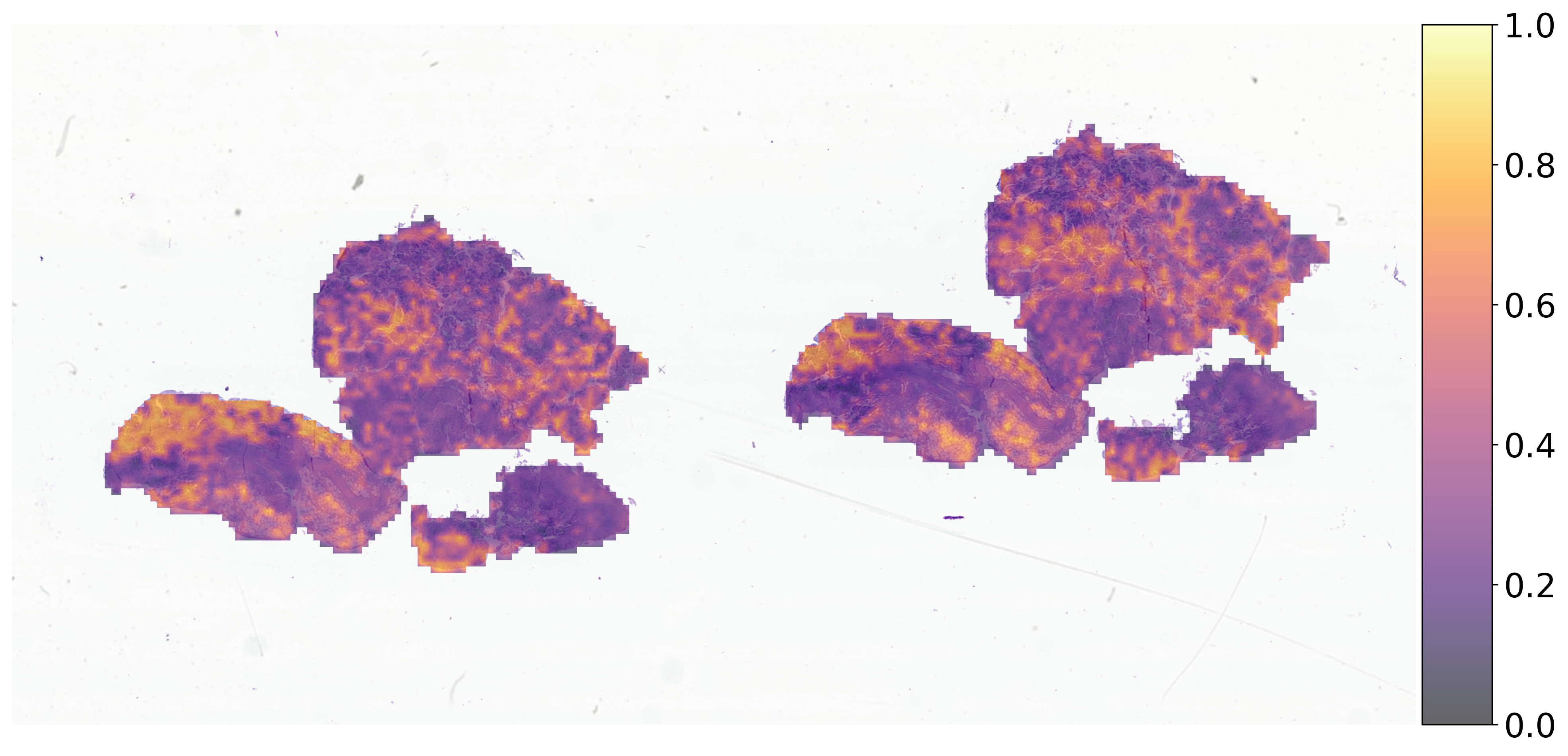}
    \caption{MIL attention map}
    \label{fig:mil_attention}
\end{subfigure}
\vspace{1em}
\begin{subfigure}[b]{0.8\linewidth}
    \centering
    \includegraphics[width=\linewidth]{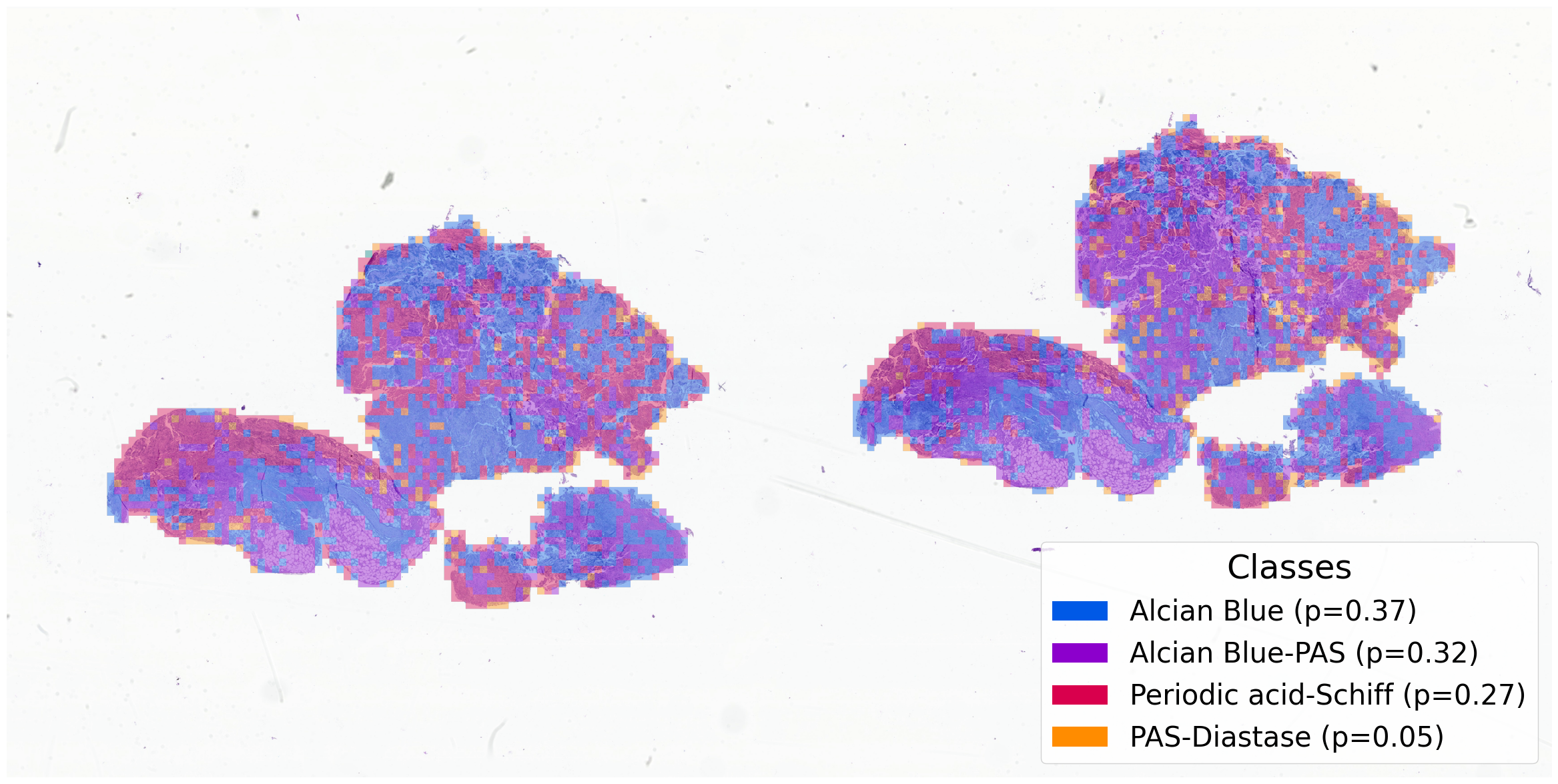}
    \caption{Patch-level predictions}
    \label{fig:voting_map}
\end{subfigure}
\caption{
Interpretability maps for an Alcian Blue–PAS stained slide.
(a) Thumbnail GradCAM.
(b) MIL attention weights map.
(c) Patch-level predictions map.
}
\label{fig:combined_maps}
\end{figure}

To qualitatively assess how the three approaches respond to a challenging Alcian Blue–PAS stained slide, we visualize model-specific interpretability maps (Fig.~\ref{fig:combined_maps}).
We compute the Grad-CAM\cite{gradcam2019} heatmap using the final token representations of the thumbnail model's ViT encoder (Fig.~\ref{fig:thumbnail_gradcam}), with the predicted class as the target.
The heatmap highlights the main tissue regions with minimal activation on background. Small tissue fragments also elicit localized responses.

For the MIL model, we visualize the attention weights assigned to individual high-resolution tiles and overlay them on the slide thumbnail (Fig.~\ref{fig:mil_attention}).
Both tissue sections show similar patterns of high-attention regions, suggesting that the MIL aggregator consistently focuses on recurring textures across the slide.

Finally, Fig.~\ref{fig:voting_map} displays the patch-level predictions for the same slide.
The predicted classes are heterogeneous: Alcian Blue (37\%) and Alcian Blue-PAS (32\%) dominate but are interspersed with PAS (27\%) and smaller proportions of PAS-D (5\%). The left tissue section has majority Alcian Blue-predicted tiles, whereas the right section contains a larger proportion of Alcian Blue-PAS predictions. This patch-level variability helps explain the difficulty of full-slide classification.

\subsection{Computational Efficiency}\label{subsection:throughput}
A critical factor for deploying quality control tools in clinical workflows is throughput. We compared the inference speeds of the three approaches. The MIL pipeline, which requires tissue segmentation, extensive patch extraction, and feature encoding, achieved a throughput of approximately $0.018$ slides per second (img/s) for $k=\text{all}$. The $k=20$ configuration improved this to $0.271$ slide/s.

The thumbnail-based approach offered the highest throughput by orders of magnitude, achieving $5.635$ slide/s. This efficiency stems from the ability to extract thumbnails directly from the pyramidal slide representation, foregoing complex pre-processing or segmentation.

\subsection{Ablation Studies}\label{subsection:ablation}
We performed ablation studies to determine the optimal input configurations for both the patch-based and thumbnail-based approaches.
All ablation studies were conducted using five-fold cross-validation on the TUM training set.
\begin{table}[ht]
\centering
\small
\begin{tabular}{l cc}
\toprule
& \multicolumn{2}{c}{\textbf{Budget}} \\
\cmidrule(lr){2-3}
\textbf{Approach} & $k{=}20$ & $k{=}\text{all}$ \\
\midrule
Voting & 0.949$\pm$0.010 & 0.953$\pm$0.011 \\[3pt]
MIL    & 0.949$\pm$0.009 & 0.955$\pm$0.011 \\
\bottomrule
\end{tabular}
\caption{Patch-aggregation approach and budget ablation study. Values represent the mean macro F1 score across 5-fold cross-validation.}
\label{tab:aggregation_ablation}
\end{table}

We compared voting, taking the mean prediction of $k$ selected patches, with Attention-based Multi Instance Learning (ABMIL), a learned aggregation of patch features.
Additionally, we compared both approaches given a fixed number of patches per slide, $k=20$ randomly selected tissue patches, and all tissue patches per slide, $k=\text{all}$.
For the fixed budget, we ran 200 rounds per slide to account for any outliers due to randomly sampling patches. We chose the highest patch magnification ($0.59$mpp) available, as the difference in number of used patches per slide is largest here between the two budgets.
Table~\ref{tab:aggregation_ablation} shows the fixed-budget constraint only minimally reduces performance for both aggregation approaches.
The difference between the approaches themselves is even smaller.
We examined the fold-wise difference between the approaches for both budgets.
The average fold-wise difference ($\text{MIL} - \text{Voting}$) in macro F1 score is $0.0006$ for the fixed budget and slightly higher for the unlimited budget ($0.0019$).
We conclude there is no meaningful difference in performance between the patch-aggregation methods for our task. As we see no benefit in pursuing two patch-aggregation methods, we focused on MIL for subsequent experiments.
\begin{figure}[t]
\centering
\begin{subfigure}[b]{0.49\linewidth}
    \centering
    \includegraphics[width=\linewidth]{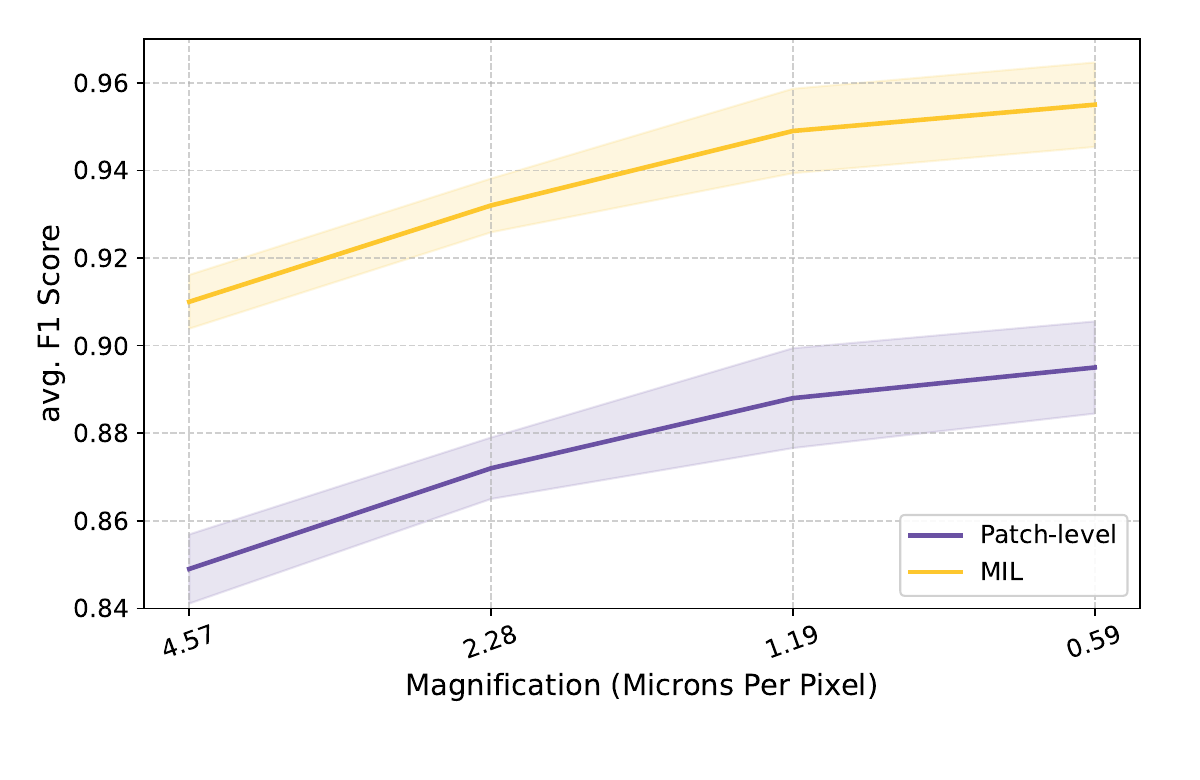}
    \caption{Patch magnification}
    \label{fig:mag_ablation}
\end{subfigure}
\hfill
\begin{subfigure}[b]{0.49\linewidth}
    \centering
    \includegraphics[width=\linewidth]{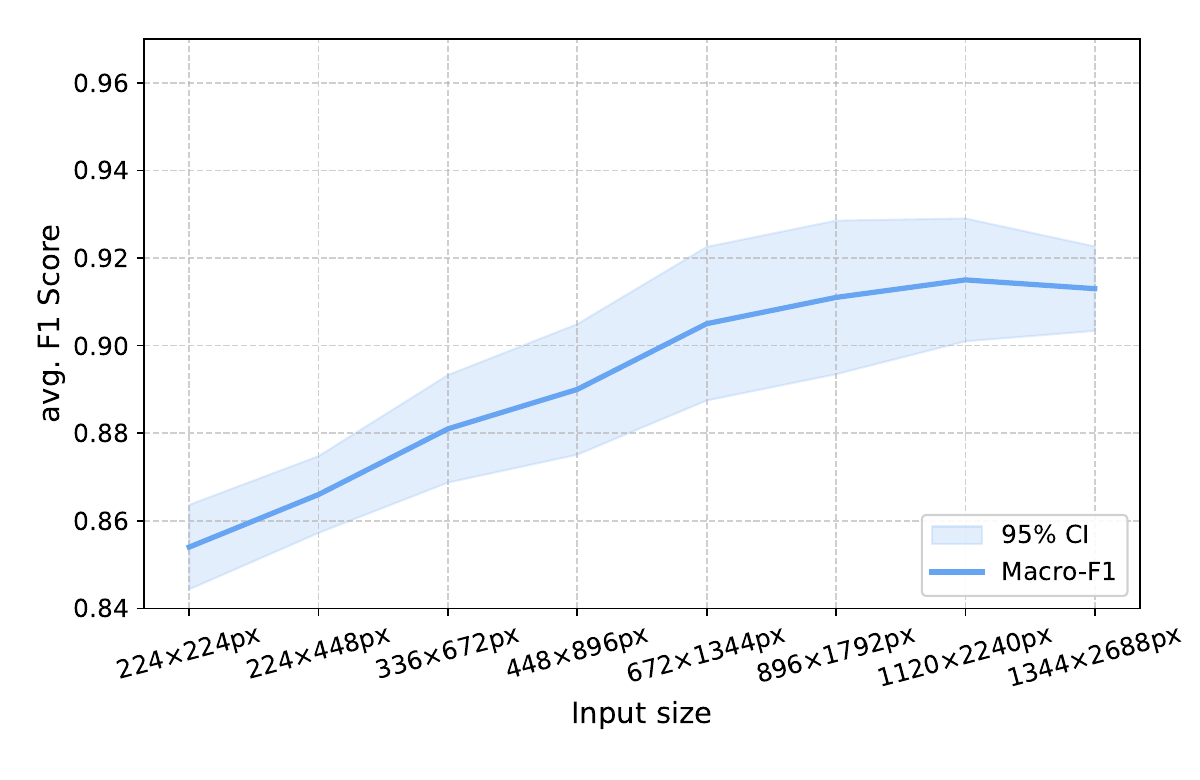}
    \caption{Thumbnail size}
    \label{fig:thumbnail_ablation}
\end{subfigure}
\caption{Ablation studies. (a)~Macro F1 score vs.\ patch magnification for patch-level and MIL ($k{=}\text{all}$) classifiers. (b)~Macro F1 score vs.\ thumbnail input resolution. Shaded areas show the 95\% CI across 5-fold cross-validation.}
\label{fig:ablation_plots}
\end{figure}

For the patch-based backbone and aggregation method, we evaluated performance across different magnifications, $4.57$, $2.28$, $1.19$ and $0.59$mpp, see Figure~\ref{fig:mag_ablation}. Both patch-level and slide-level classification benefited from higher magnifications, with the highest F1 scores achieved at the highest magnification, $0.59$mpp ($0.879 \pm 0.018$ patch-level macro F1 score; $0.939 \pm 0.017$ MIL slide-level macro F1 score). Consequently, $0.59$mpp magnification was selected for subsequent experiments.

For the thumbnail-based approach, we investigated the impact of input resolution on classification performance. Figure~\ref{fig:thumbnail_ablation} shows increasing the thumbnail resolution steadily improved the macro F1 score.
Due to the increased computational cost of processing larger inputs yielding negligible returns, we standardized on $896 \times 1792$ px for the thumbnail classifier.

\section{Discussion}\label{section:discussion}

Our results show that a lightweight thumbnail-based classifier can achieve performance that is competitive with computationally intensive patch-based methods on internal data, while providing markedly better generalization to external cohorts and substantially higher throughput. Although the Multi-Instance Learning (MIL, $k=\text{all}$) approach achieved the highest macro F1 score on the internal test set, its advantage over the thumbnail classifier diminished considerably when closely related stains were merged into coarse categories. This pattern suggests that the main strengths of MIL arise from its ability to exploit fine-grained, high-magnification textures that differentiate closely related stains (e.g., PAS vs. PAS-D), whereas the global structural cues present in downsampled thumbnails are already sufficient for the majority of distinct staining types.

Across ablation studies, we found that the benefit of increasing patch counts diminishes rapidly. A fixed budget of 20 randomly sampled patches was nearly indistinguishable from using all available tissue patches for MIL, both in cross-validation and on the internal test set. This indicates that a small, diverse subset of high-magnification regions captures most of the discriminatory signal present in the full slide, reinforcing our findings that slide-level stain classification relies on a combination of dominant global cues and a limited number of highly informative local features.

On the external TCGA dataset, the thumbnail approach generalized better than all MIL configurations. Despite being the least accurate on the internal test slides, the thumbnail classifier outperformed MIL at both budgets for H\&E-FFPE/H\&E-FS differentiation. This advantage persisted when training a dedicated binary fixation type predictor, where thumbnails achieved high macro F1 and AUROC values. We hypothesize that patch-based methods are more susceptible to overfitting to institute-specific patterns, such as local staining intensities, nuclear contrast, and scanner-dependent high-frequency artefacts. In contrast, the heavily downscaled thumbnail representation suppresses these local idiosyncrasies and encourages the model to rely on coarser, more transferable global morphology and color composition. This inductive bias appears to be beneficial for domain generalization under realistic cross-laboratory shifts in staining and scanning protocols.

The interpretability analyses further support these differences. MIL attention maps focused on recurring high-resolution textures that were specific to individual stains, while Grad-CAM on thumbnails highlighted broad tissue regions with little sensitivity to small artefacts. Patch-level prediction maps for challenging stains revealed substantial heterogeneity within individual slides, suggesting that patch aggregation methods integrate a more complex mixture of local signals. These qualitative differences provide a mechanistic explanation for why patch-based models excel at fine-grained tasks yet can generalize poorly when local textures shift across laboratories.

From a deployment perspective, throughput is a central consideration for automated quality control. The MIL pipeline is computationally demanding due to the reliance on tissue segmentation, patch extraction, and feature encoding. Even with a fixed patch budget, its throughput remains well below 1 slide per second. The thumbnail classifier, in contrast, benefits from direct extraction of low-resolution pyramid layers and achieves up to two orders of magnitude higher throughput than MIL, enabling near real-time analysis in high-volume clinical archives. In practical QC settings where thousands of slides may be ingested daily, this computational efficiency is essential.

This study has several limitations. The models were trained and evaluated primarily on data from a single institution, and histological stains are known to exhibit substantial inter-laboratory variability due to differences in reagents, counterstains, and scanner hardware. We therefore do not expect a single universal model to perform reliably across all sites without adaptation. Rather than enforcing a closed set of stain categories, future systems may benefit from open-set or out-of-distribution detection to flag slides whose staining characteristics deviate from the training distribution. Uncertainty estimation approaches such as Bayesian neural networks\cite{BNN2020, BNN_Goan_2020} or Monte Carlo dropout\cite{montecarloDropout2015} could provide reliable mechanisms to identify ambiguous or previously unseen staining variants and route them for manual inspection. In addition, integrating metadata consistency checks, comparing model predictions with case-level information when available, offers a simple but effective safeguard. Our performance analysis focused on prediction throughput, not taking into account computational resource and energy consumption. Future work could explore more efficient model architectures for the thumbnail-based approach to reduce GPU memory needs.

Overall, our findings indicate that while patch-based methods remain the gold standard for maximum slide-level precision, thumbnail-based classification provides a favourable balance of accuracy, robustness, and computational efficiency. This balance makes thumbnail classifiers particularly well suited for scalable, institute-level quality control systems, where high throughput and resilience to domain shift are critical for practical adoption.

\section{Methods}\label{section:methods}

\subsection{Datasets and Study Population}\label{subsection:data}

\begin{figure}[ht]
\centering
\begin{subfigure}[b]{0.47\linewidth}
\centering
\includegraphics[width=\linewidth]{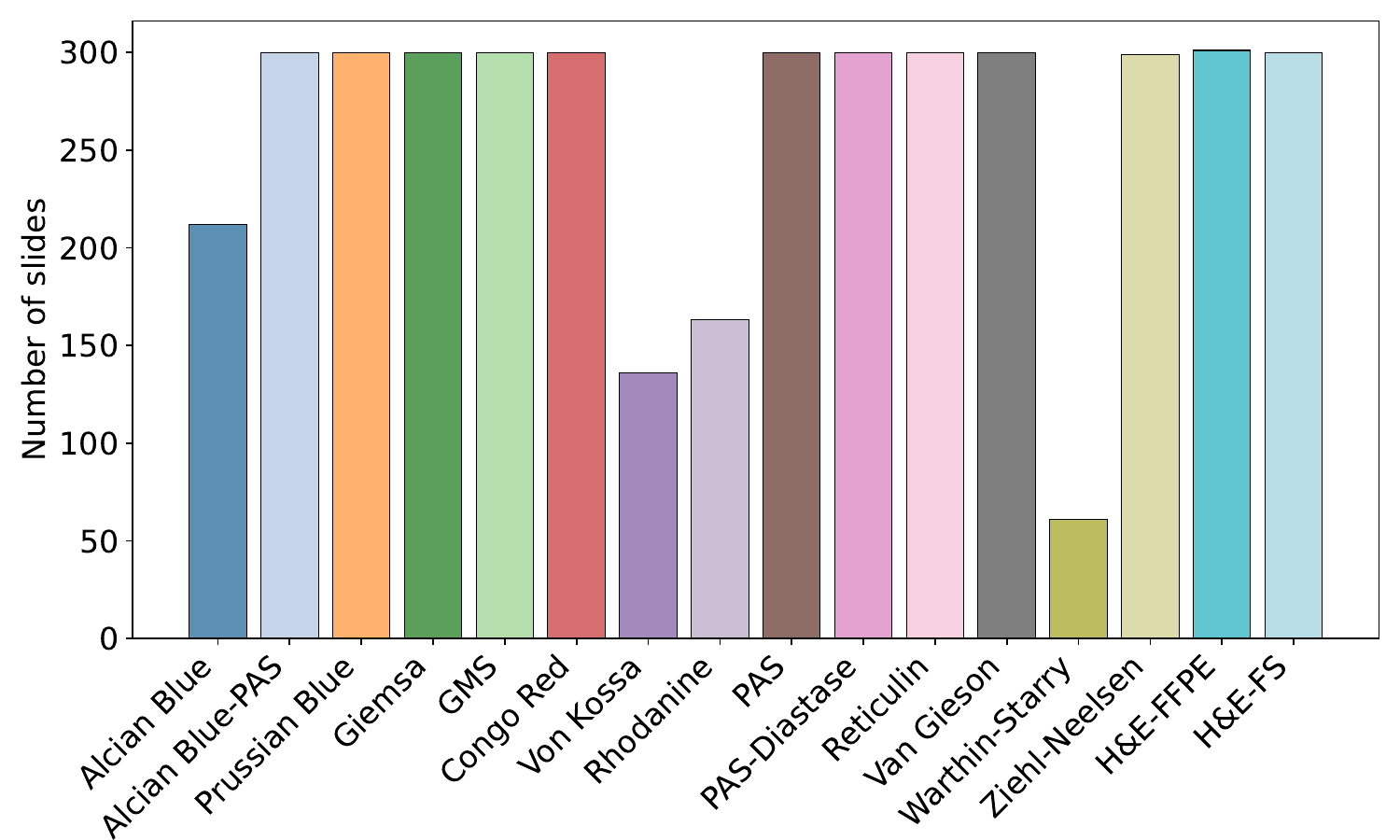}
\caption{Distribution of stains in the TUM dataset.}
\label{fig:stain_distribution}
\end{subfigure}
\begin{subfigure}[b]{0.49\linewidth}
\centering
\includegraphics[width=\linewidth]{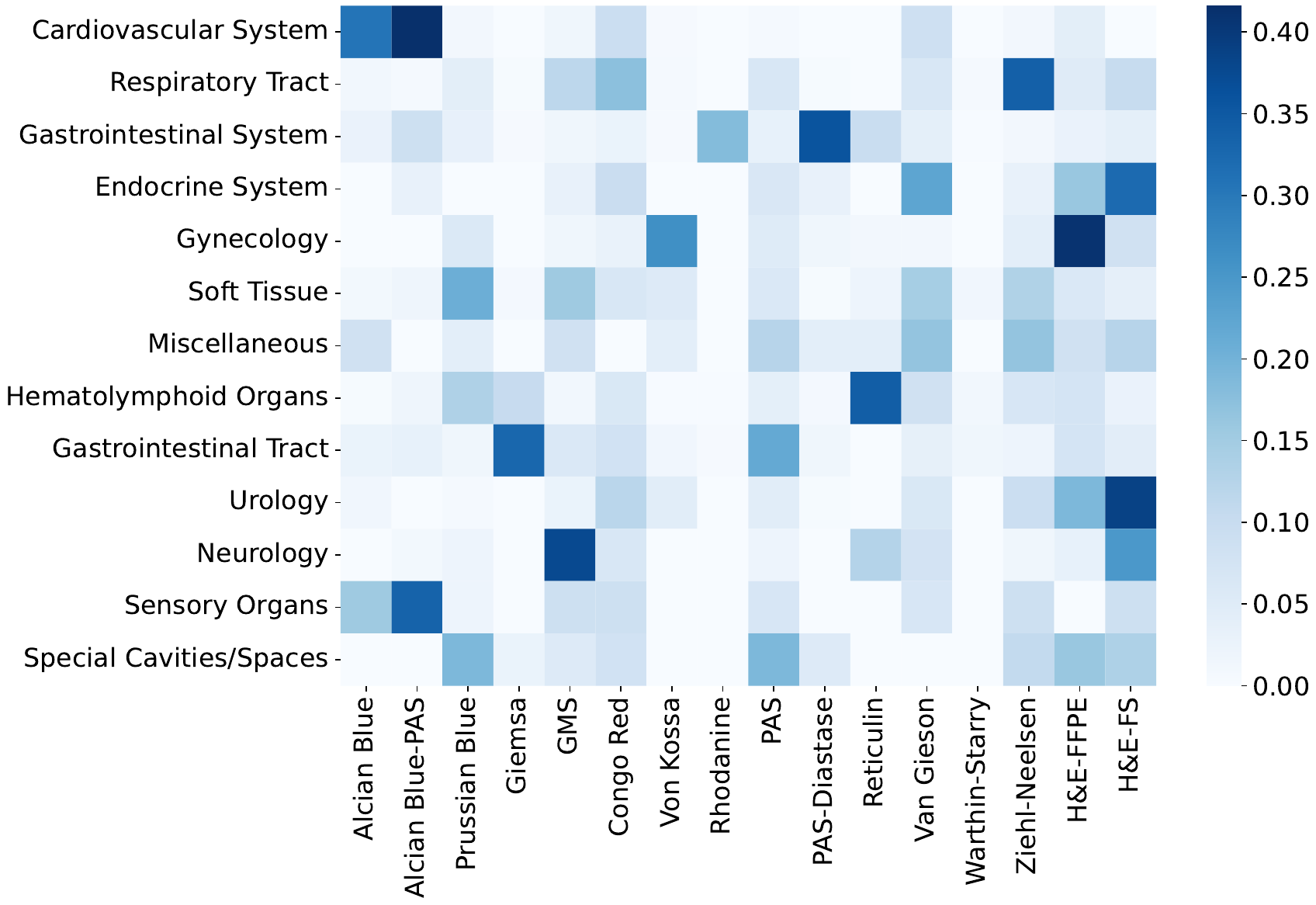}
\caption{Co-occurrence matrix of Locality and Stain in TUM dataset. Locality is an anatomical grouping of organ codes.}
\label{fig:locality_stain_co-occurrence}
\end{subfigure}
\caption{Stain distribution and co-occurrence with anatomical locality in the TUM dataset.}
\end{figure}

\begin{figure}[h]
\centering

\begin{subfigure}[b]{0.25\linewidth}
    \includegraphics[width=\linewidth]{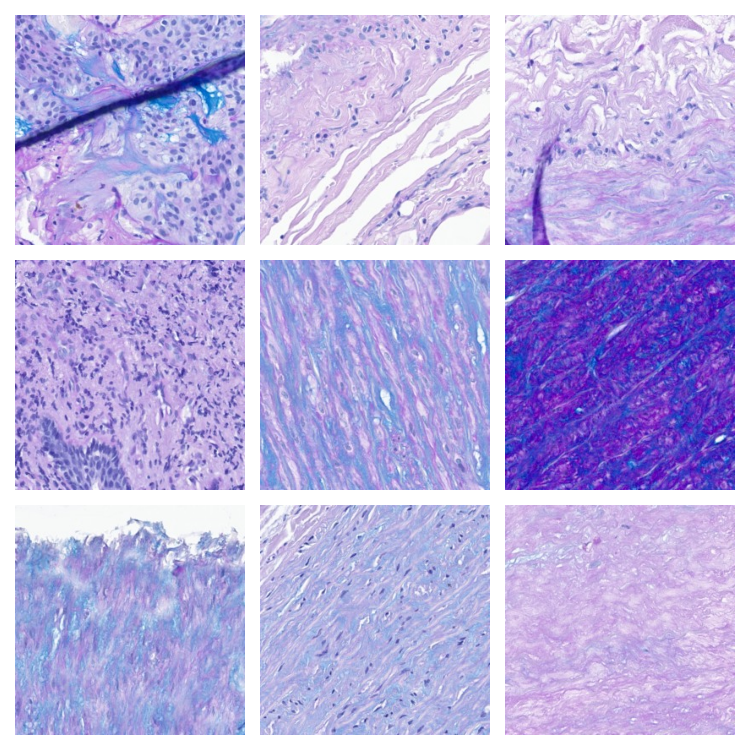}
    \caption{Alcian Blue}
\end{subfigure}
\begin{subfigure}[b]{0.25\linewidth}
    \includegraphics[width=\linewidth]{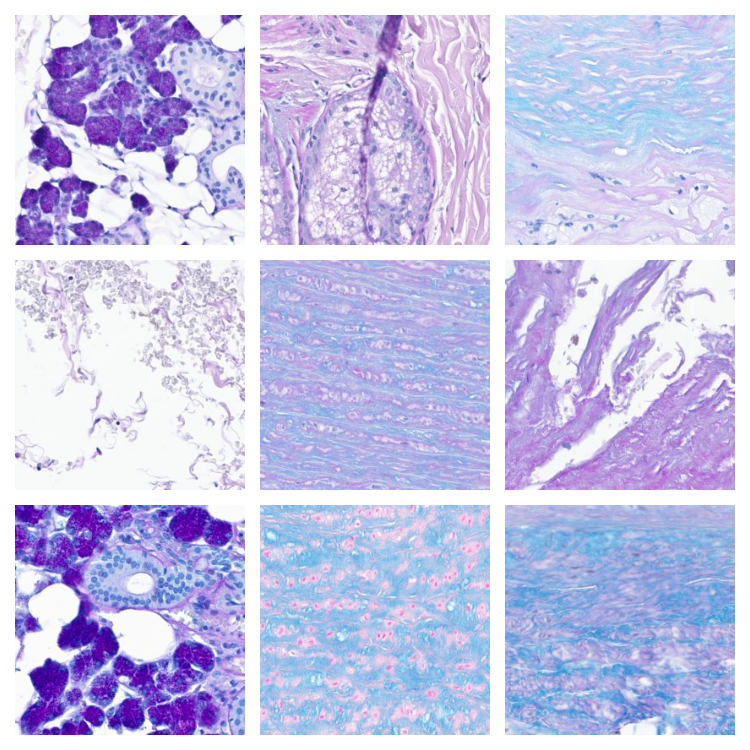}
    \caption{Alcian Blue–PAS}
\end{subfigure}

\vspace{1em}

\begin{subfigure}[b]{0.25\linewidth}
    \includegraphics[width=\linewidth]{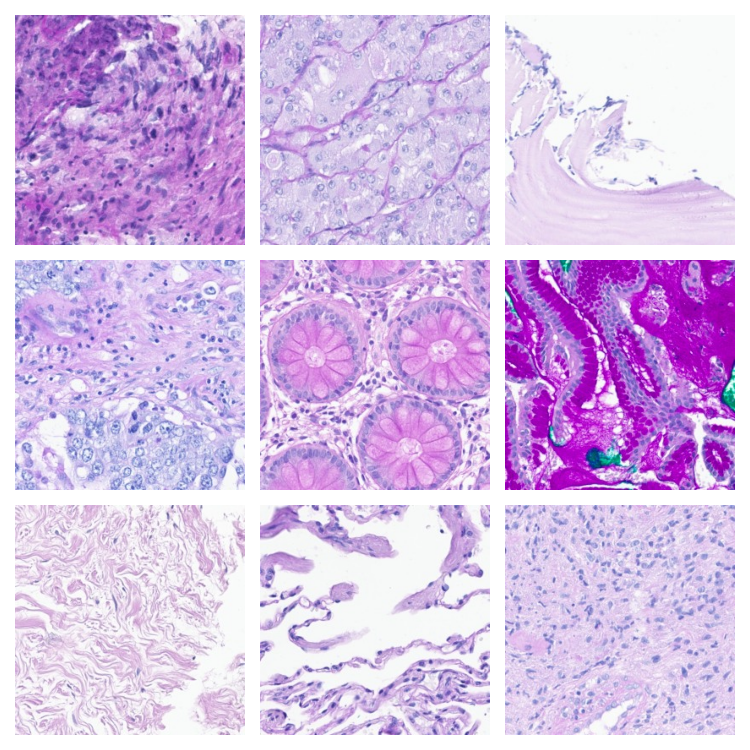}
    \caption{PAS}
\end{subfigure}
\begin{subfigure}[b]{0.25\linewidth}
    \includegraphics[width=\linewidth]{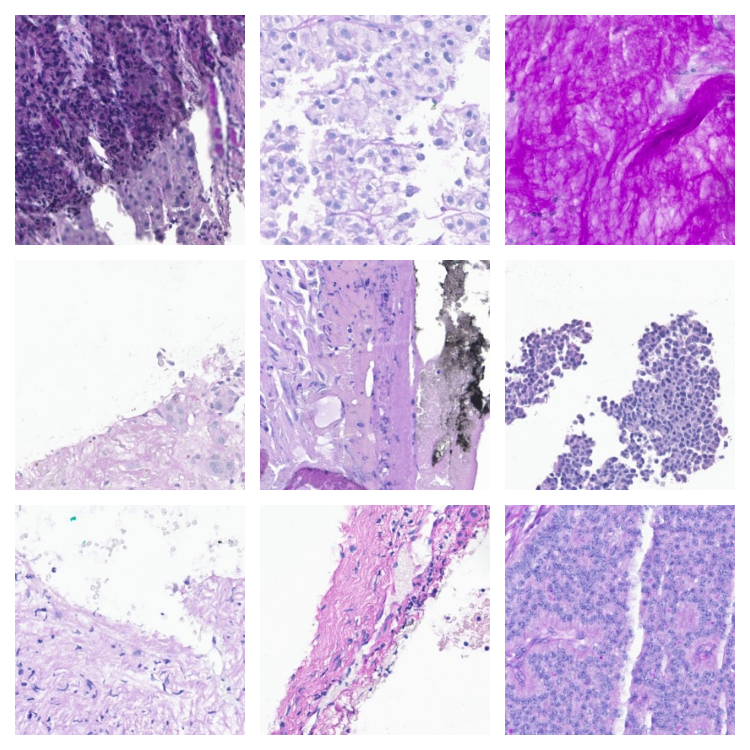}
    \caption{PAS–D}
\end{subfigure}

\caption{
Representative high-magnification tiles illustrating the visual similarity between stains merged in the coarse class set.
Alcian Blue vs.\ Alcian Blue–PAS and PAS vs.\ PAS–D exhibit only subtle differences, motivating the merged \emph{coarse} class definitions.
}
\label{fig:stain_examples}
\end{figure}

We compiled a dataset of whole slide images (WSIs) representing the 14 most common special stains used at the Institute of Pathology, Technical University of Munich (TUM), alongside standard Haematoxylin and Eosin (H\&E-FFPE) and frozen section H\&E (H\&E-FS). The complete dataset consists of 4,172 slides. We collected a maximum of 300 samples per stain class. During data collection, a mislabeled slide (H\&E-FFPE erroneously annotated as Ziehl-Neelsen) was identified and corrected, resulting in a final count of 301 H\&E-FFPE slides.
Figure~\ref{fig:stain_distribution} shows the distribution of stains in our TUM dataset. While most stains are represented with 300 slides, several stains are comparatively rare. In particular, Alcian Blue has 212 samples, Rhodanine 163 and Von Kossa 136. The rarest stain is Warthin--Starry, with only 61 samples. We report results for a ``fine'' (all 16 stains) and a ``coarse'' class set, where visually similar stain pairs were merged (Figure~\ref{fig:stain_examples}).

We show the co-occurrence of stains with anatomical ``localities'' in the TUM dataset in Figure~\ref{fig:locality_stain_co-occurrence}. Here, locality denotes a coarse anatomical grouping of the routine organ codes used in our laboratory information system. Rows correspond to localities, columns to stains, and colour intensity reflects the relative frequency with which a stain is used in a given locality. The heatmap highlights that some stains are strongly enriched in specific anatomical regions, whereas routine stains such as H\&E-FFPE and H\&E-FS are used across almost all localities.

For internal validation, we employed a 5-fold cross-validation scheme. We also reserved 20\% of each class as a held-out test set.
For external testing, we utilized data from The Cancer Genome Atlas (TCGA) to assess the model's ability to differentiate between H\&E-FFPE and H\&E-FS in a multi-center setting.

\subsection{Image Pre-processing}\label{subsection:preprocessing}

Tissue regions were segmented from the background using HEST\cite{HEST2024} via the TRIDENT\cite{TRIDENT2025} framework. Due to the lack of segmentation models trained specifically for special stains, all segmentations underwent a manual review process to correct egregious errors.

For the patch-based approaches (MIL and voting), the segmented tissue regions were tessellated into patches. 
We extracted patches at multiple resolutions to determine the optimal magnification, ultimately selecting 40x magnification with a patch size of $512 \times 512$px ($0.59$ mpp) for the final models.

For the thumbnail-based approach, slide thumbnails were extracted directly using the OpenSlide library\cite{openslide} by downscaling the nearest available pyramid level to $1792 \times 3584$.
Images were rotated before downscaling so that the width was the longer side.
We evaluated effective thumbnail resolutions ranging from $224 \times 224$px to $1344 \times 2688$px, standardized on $896 \times 1792$px ($17.99$ avg. mpp) for the final model.

\subsection{Model Architecture}\label{subsection:model}

All methods use the H0-mini foundation model as the visual backbone.
H0-mini\cite{h0mini} is a distilled ViT-Base\cite{transformer2017, vit2020} (86M parameters, patch size 14, 768-dimensional embeddings), trained via knowledge distillation from the larger H-Optimus-0\cite{hoptimus0} (ViT-g/14) model, which itself was pre-trained using DINOv2\cite{DINOv2} on over 500,000 H\&E slides.

On top of this backbone, we employed three distinct slide-level architectures:
\begin{itemize}
    \item Patch-level classifier: A single linear layer operating on the CLS token concatenated with the mean patch token. Used to train the backbone in a supervised fashion. Predictions were used for the voting approach.
    \item Multiple Instance Learning (MIL): An Attention-Based MIL (ABMIL\cite{ABMIL2018}) network consisting of: patch embedder, gated attention module, small MLP classifier.
    \item Thumbnail classifier: An H0-mini encoder configured for larger 2D inputs (up to $896 \times 1792$px) followed by a two-layer MLP head (512 and 64 units with PReLU, BatchNorm, and dropout\cite{dropout2012}).
\end{itemize}

\subsection{Classification Pipelines}\label{subsection:pipelines}

We evaluated three pipelines: thumbnail, voting, and MIL. Each pipeline uses the H0-mini backbone differently.

The thumbnail method processes the entire slide thumbnail in a single forward pass (Figure~\ref{fig:fig_1_thumbnail}).
H0-mini was adapted via timm\cite{timm2019} to accept large-resolution thumbnails ($896 \times 1792$px).
Training proceeds in two steps:
\begin{itemize}
    \item Patch-level fine-tuning: Fine-tune H0-mini on $224 \times 224$px patches extracted at $4.57$ mpp.
    \item Thumbnail-level fine-tuning: Initialize the thumbnail model with the patch-level weights and fine-tune end-to-end on full-resolution thumbnails.
\end{itemize}

The voting pipeline aggregates independent predictions from high-magnification patches.
\begin{itemize}
    \item Fine-tuning: H0-mini is fine-tuned on $224 \times 224$px tissue patches (0.59 mpp) using a linear classifier head.
    \item Aggregation: Slide-level predictions are computed as the mean probability across $k$ randomly sampled patches.
\end{itemize}
We explored $k \in [10, 1000]$ and selected $k=20$ as the efficient configuration.
The $k=\text{all}$ variant aggregates all available tissue patches.

MIL aggregates patch-level embeddings into a single representation (Figure~\ref{fig:fig_1_mil}).
\begin{itemize}
    \item Feature extraction: Patch embeddings are generated using the same fine-tuned H0-mini encoder as in the voting pipeline. Features are computed for all tissue patches or for a random subset of size $k$.
    \item MIL aggregation: A gated attention-based MIL network\cite{ABMIL2018} is trained to assign patch-level attention weights, aggregate into a slide-level representation and arrive at a final slide-level prediction.
\end{itemize}
During MIL training, the H0-mini encoder remains frozen. Both $k=20$ and $k=\text{all}$ configurations were evaluated.

\subsection{Fine-Tuning}\label{subsection:fine-tuning}

All training used cross-entropy loss\cite{pytorch_celoss} with $0.05$ label smoothing\cite{label_smoothing2019}, as well as cosine annealing with warm-up.
We used early-stopping after five epochs of no increase (> 0.001) in validation macro F1 score for the ablation studies.

Patch-level fine-tuning:
\begin{itemize}
    \item Optimizer: AdamW\cite{pytorch_adamw} (lr$=5\times10^{-5}$, weight decay$=0.01$).
    \item Dropout, drop-path, and attention-drop for H0-mini: $0.5$.
    \item Batch size: $256$.
    \item Epochs: $15$.
    \item Encoder output tokens: CLS + mean patch token.
\end{itemize}

These weights initialize the thumbnail model and serve as the feature extractor for voting and MIL.

MIL training:
\begin{itemize}
    \item Optimizer: AdamW (lr$=1\times10^{-5}$, weight decay$=5\times10^{-2}$).
    \item Batch size: $256$.
    \item Epochs: $15$.
    \item Only the MIL module is trainable; H0-mini is frozen.
\end{itemize}

Thumbnail training:
\begin{itemize}
    \item Optimizer: 8-bit AdamW\cite{torchao} (lr$=1\times10^{-5}$, weight decay$=5\times10^{-2}$).
    \item Batch size: $128$.
    \item Epochs: $20$.
    \item Backbone: fully unfrozen H0-mini (dropout, drop-path, and attention-drop $0.3$); only CLS token.
\end{itemize}

Fixation-type prediction (binary):
For the H\&E-FFPE vs. H\&E-FS task, the same thumbnail and MIL training configurations were used, only changing the final classifier layer. No dedicated patch-level fine-tuning was performed.

\subsection{Latency Analysis}\label{subsection:latency_setup}
Throughput was measured as the average number of whole slide images processed in one second (img/s). For each pipeline, timing captured the effective per-slide processing time, defined as the wall-clock duration from slide loading to completion of all required preprocessing and model inference steps, while excluding one-off model-loading overheads and optional visualization routines.

For the MIL pipelines, the measured time included slide initialization, tissue segmentation using TRIDENT, tissue-coordinate extraction, and patch-feature encoding.
For the thumbnail approach, the measured time included slide initialization, thumbnail extraction, preprocessing, and forward inference through the thumbnail classifier, excluding checkpoint loading and model construction.

All runtime measurements were performed on a TCGA slide containing 8,246 tissue patches, close to the dataset-wide average of 8,245 patches per slide. Each measurement was repeated 25 times.
We conducted the measurements on an Nvidia GraceHopper server with a single Nvidia H100 96GB, 72 ARM CPU-cores and 480GB RAM.

\bibliographystyle{unsrt}
\bibliography{sample}

\section*{Acknowledgements}

The results published here are in part based upon data generated by the TCGA Research Network: https://www.cancer.gov/tcga.

\end{document}